# Acceleration of Deep Neural Network Training with Resistive Cross-Point Devices

**Authors:** Tayfun Gokmen,* Yurii Vlasov

**Affiliations**
IBM T.J. Watson Research Center, Yorktown Heights, NY 10598 USA
*Correspondence to: tgokmen@us.ibm.com

**Abstract**
In recent years, deep neural networks (DNN) have demonstrated significant business impact in large scale analysis and classification tasks such as speech recognition, visual object detection, pattern extraction, etc. Training of large DNNs, however, is universally considered as time consuming and computationally intensive task that demands datacenter-scale computational resources recruited for many days. Here we propose a concept of resistive processing unit (RPU) devices that can potentially accelerate DNN training by orders of magnitude while using much less power. The proposed RPU device can store and update the weight values locally thus minimizing data movement during training and allowing to fully exploit the locality and the parallelism of the training algorithm. We identify the RPU device and system specifications for implementation of an accelerator chip for DNN training in a realistic CMOS-compatible technology. For large DNNs with about 1 billion weights this massively parallel RPU architecture can achieve acceleration factors of $30,000X$ compared to state-of-the-art microprocessors while providing power efficiency of $84,000\ GigaOps/s/W$. Problems that currently require days of training on a datacenter-size cluster with thousands of machines can be addressed within hours on a single RPU accelerator. A system consisted of a cluster of RPU accelerators will be able to tackle Big Data problems with trillions of parameters that is impossible to address today like, for example, natural speech recognition and translation between all world languages, real-time analytics on large streams of business and scientific data, integration and analysis of multimodal sensory data flows from massive number of IoT (Internet of Things) sensors.

MAIN TEXT

1. Introduction

Deep Neural Networks (DNNs) [1] demonstrated significant commercial success in the last years with performance exceeding sophisticated prior methods in speech [2] and object recognition [3–5]. However, training the DNNs is an extremely computationally intensive task that requires massive computational resources and enormous training time that hinders their further application. For example, a 70% relative improvement has been demonstrated for a DNN with 1 billion connections that was trained on a cluster with 1000 machines for three days [6]. Training the DNNs relies in general on the backpropagation algorithm that is intrinsically local and parallel [7]. Various hardware approaches to accelerate DNN training that are exploiting this locality and parallelism have been explored with a different level of success starting from the early 90s [8,9] to current developments with GPU [10,11], FPGA [12] or specially designed ASIC [13]. Further acceleration is possible by fully utilizing the locality and parallelism of the algorithm. For a fully connected DNN layer that maps $N$ neurons to $N$ neurons significant acceleration can be achieved by minimizing data movement using local storage and processing of the weight values on the same node and connecting nodes together into a massive $N \times N$ systolic array [8] where the whole DNN can fit in. Instead of a usual time complexity of $O(N^2)$ the problem can be reduced therefore to a constant time $O(1)$ independent of the array size. However, the addressable problem size is limited to the number of nodes in the array that is challenging to scale up to billions even with the most advanced CMOS technologies.

Novel nano-electronic device concepts based on non-volatile memory (NVM) technologies, such as phase change memory (PCM) [14,15] and resistive random access memory (RRAM) [15–19], have been explored recently for implementing neural networks with a learning rule inspired by spike-timing-dependent plasticity (STDP) observed in biological systems [20]. Only recently, their implementation for acceleration of DNN training using backpropagation algorithm have been considered [21–25] with reported acceleration factors ranging from 27X [26] to 900X [21], and even 2140X [27] and significant reduction in power and area. All of these bottom-up approach of using previously developed memory technologies looks very promising, however the estimated acceleration factors are limited by device specifications intrinsic to their application as NVM cells. Device characteristics usually considered beneficial or irrelevant for memory applications such as high on/off ratio, digital bit-wise storage, and asymmetrical set and reset operations, are becoming limitations for acceleration of DNN training [26,28]. These non-ideal device characteristics can potentially be compensated with a proper design of peripheral circuits and a whole system, but only partially and with a cost of significantly increased operational time [26]. In contrast, here we propose an up-down approach where ultimate acceleration of DNN training is achieved by design of a system and CMOS circuitry that imposes specific requirements for resistive devices. We propose and analyze a concept of Resistive Processing Unit (RPU) devices that can simultaneously store and process weights and are potentially scalable to billions of nodes with foundry CMOS technologies. Our estimates indicate that acceleration factors close to $30,000X$ are achievable on a single chip with realistic power and area constraints.

## 2. Definition of the RPU device concept

The backpropagation algorithm is composed of three cycles, forward, backward and weight update that are repeated many times until a convergence criterion is met. The forward and backward cycles mainly involve computing vector-matrix multiplication in forward and backward directions. This operation can be performed on a 2D crossbar array of two-terminal resistive devices as it was proposed more than 50 years ago [29]. In forward cycle, stored conductance values in the crossbar array form a matrix, whereas the input vector is transmitted as voltage pulses through each of the input rows. In a backward cycle, when voltage pulses are supplied from columns as an input, then the vector-matrix product is computed on the transpose of a matrix. These operations achieve the required $O(1)$ time complexity, but only for two out of three cycles of the training algorithm.

In contrast to forward and backward cycles, implementing the weight update on a 2D crossbar array of resistive devices locally and all in parallel, independent of the array size, is challenging. It requires calculating a vector-vector outer product which consist of a multiplication operation and an incremental weight update to be performed locally at each cross-point as illustrated in Fig. 1A. The corresponding update rule is usually expressed as [7]

$$w_{ij} \leftarrow w_{ij} + \eta x_i \delta_j \tag{1}$$

where $w_{ij}$ represents the weight value for the $i^{th}$ row and the $j^{th}$ column (for simplicity layer index is omitted) and $x_i$ is the activity at the input neuron, $\delta_j$ is the error computed by the output neuron and $\eta$ is the global learning rate.

In order to implement a **local and parallel** update on an array of two-terminal devices that can perform both weight storage and processing (Resistive Processing Unit or RPU) we first propose to significantly simplify the multiplication operation itself by using stochastic computing techniques [30–32]. It has been shown that by using two stochastic streams the multiplication operation can be reduced to a simple AND operation [30–32]. Fig. 1B illustrates the stochastic update rule where numbers that are encoded from neurons ($x_i$ and $\delta_j$) are translated to stochastic bit streams using stochastic translators (STR). Then they are sent to the crossbar array where each RPU device changes its conductance ($g_{ij}$) slightly when bits from $x_i$ and $\delta_j$ coincide. In this scheme we can write the update rule as follows

$$w_{ij} \leftarrow w_{ij} \pm \Delta w_{min} \sum_{n=1}^{BL} A_i^n \wedge B_j^n \tag{2}$$

where $BL$ is length of the stochastic bit stream at the output of STRs that is used during the update cycle, $\Delta w_{min}$ is the change in the weight value due to a single coincidence event, $A_i^n$ and $B_j^n$ are random variables that are characterized by Bernoulli process, and a superscript $n$ represents bit position in the trial sequence. The probabilities that $A_i^n$ and $B_j^n$ are equal to unity are controlled by $Cx_i$ and $C\delta_j$, respectively, where $C$ is a gain factor in the STR.

One possible pulsing scheme that enables the stochastic update rule of Eq.2 is presented in Fig. 1C. The voltage pulses with positive and negative amplitudes are sent from corresponding STRs on rows ($A_i$) and columns ($B_j$), respectively. As opposed to a floating point number encoded into a binary stream, the corresponding number translated into a stochastic stream is represented by a whole population of such pulses. In order for a two-terminal RPU device to distinguish coincidence events at a cross-point, its conductance value should not change significantly when a single pulse $V_S/2$ is applied to a device from a row or a column. However, when two pulses coincide and the RPU device sees the full voltage ($V_S$) the conductance should change by nonzero amount $\Delta g_{min}$. The parameter $\Delta g_{min}$ is proportional to $\Delta w_{min}$ through the amplification factor defined by peripheral circuitry. To enable both up and down changes in conductance the polarity of the pulses can be switched during the update cycle as shown in Fig. 1D. The proposed pulsing scheme allows all the RPU devices in an array to work in parallel and perform the multiplication operation locally by simply relying on the statistics of the coincidence events, thus achieving the $O(1)$ time complexity for the weight update cycle of the training algorithm.

3. **Network training with RPU array using stochastic update rule**

To test the validity of this approach, we compare classification accuracies achieved with a deep neural network composed of fully connected layers with 784, 256, 128 and 10 neurons, respectively. This network is trained with a standard MNIST training dataset of 60,000 examples of images of handwritten digits [33] using cross-entropy objective function and backpropagation algorithm [7]. Raw pixel values of each 28x28 pixel image are given as inputs, while sigmoid and softmax activation functions are used in hidden and output layers, respectively. The temperature parameter for both activation functions is assumed to be unity. Fig. 2 shows a set of classification error curves for the MNIST test dataset of 10,000 images. The curve marked with open circles in Fig. 2A corresponds to a baseline model where the network is trained using the conventional update rule as defined by Eq.1 with a floating point multiplication operation. Typically, batch training is performed to decrease the number of updates and hence reduce the overall training time. Here, in order to test the most update demanding case, the batch size of unity is chosen throughout the following experiments. Training is performed repeatedly for all 60,000 images in training dataset which constitutes a single training epoch. Learning rates of $\eta = 0.01$, 0.005 and 0.0025 for epochs 0-10, 11-20 and 21-30, respectively, are used. The baseline model reaches classification error of 2.0% on the test data in 30 epochs.

In order to make a fair comparison between the baseline model and the stochastic model in which the training uses the stochastic update rule of Eq.2, the learning rates need to match. In the most general form the average change in the weight value for the stochastic model can be written as

$$\mathbb{E}(\Delta w_{ij}) = BL\,\Delta w_{min}\,C^2\,x_i \delta_j \tag{3}$$

Therefore the learning rate for the stochastic model is controlled by three parameters $BL$, $\Delta w_{min}$, and $C$ that should be adjusted to match the learning rates that are used in the baseline model.

Although the stochastic update rule allows to substitute multiplication operation with a simple AND operation, the result of the operation, however, is no longer exact, but probabilistic with a standard deviation to mean ratio that scales with $1/\sqrt{BL}$. Increasing the stochastic bit stream length $BL$ would decrease the error, but in turn would increase the update time. In order to find an acceptable range of $BL$ values that allow to reach classification errors similar to the baseline model, we performed training using different $BL$ values while setting $\Delta w_{min} = \eta/BL$ and $C = 1$ in order to match the learning rates used for the baseline model as discussed above. As it is shown in Fig. 2A, $BL$ as small as 10 is sufficient for the stochastic model to become indistinguishable from the baseline model.

To determine how strong non-linearity in the device switching characteristics is required for the algorithm to converge to classification errors comparable to the baseline model, a non-linearity factor is varied as shown Fig. 2B. The non-linearity factor is defined as the ratio of two conductance changes at half and full voltages as $k = \frac{\Delta g(V_S/2)}{\Delta g(V_S)}$. As shown in Fig. 2C, the values of $k \approx 1$ correspond to a saturating type non-linear response, when $k = 0.5$ the response is linear as typically considered for a memristor [34], and values of $k \approx 0$ corresponds to a rectifying type non-linear response. As it is shown in Fig. 2B the algorithm fails to converge for the linear response, however, a non-linearity factor $k$ below 0.1 is enough to achieve classification errors comparable to the baseline model.

These results validate that although the updates in the stochastic model are probabilistic, classification errors can become indistinguishable from those achieved with the baseline model. The implementation of the stochastic update rule on an array of analog RPU devices with non-linear switching characteristics effectively utilizes the locality and the parallelism of the algorithm. As a result the update time is becoming independent of the array size, and is a constant value proportional to $BL$, thus achieving the required $O(1)$ time complexity.

### 4. Derivation of RPU device specifications

Various materials, physical mechanisms, and device concepts have been analyzed in view of their potential implementation as cross-bar arrays for neural network training [21–26]. These technologies have been initially developed for storage class memory applications. It is not clear beforehand, however, whether intrinsic limitations of these technologies, when applied to realization of the proposed RPU concept, would result in a significant acceleration, or, in contrast, might limit the performance. For example, PCM devices can only increase the conductance during training, thus resulting in network saturation after a number of updates. This problem can be mitigated by a periodic serial reset of weights, however with a price of lengthening the training time [22,26] as it violates the $O(1)$ time complexity. In order to determine the device specifications required to achieve the ultimate acceleration when $O(1)$ time complexity is reached, we performed a series of trainings summarized in Fig. 3. Each figure corresponds to a specific "stress test" where a single parameter is scanned while all the others are fixed allowing to explore the acceptable RPU device parameters that the algorithm can tolerate without significant error penalty. This includes variations in RPU device switching characteristics, such as, incremental conductance change due to a single coincidence event, asymmetry in up and down conductance changes, tunable range of the conductance values, and various types of noise in the system. For all of the stochastic models illustrated in Fig. 3, $k = 0$ and $BL = 10$ is used. In order to match the learning rates used for the baseline model the $x_i$ and $\delta_j$ are translated to stochastic streams with $C$ defined as $C = \sqrt{\eta/(BL\,\Delta w_{min})}$. This allows the average learning rate to be the same as in the baseline model.

Ideally, the RPU device should be analog i.e. the conductance change due to a single coincidence event $\Delta g_{min}$ should be arbitrarily small, thus continuously covering all the allowed conductance values. To determine the largest acceptable $\Delta g_{min}$ due to a single coincidence event that does not produce significant error penalty, the parameter $\Delta w_{min}$ is scanned between 0.32 and 0.00032, while other parameters are fixed as shown in Fig. 3A. While for large $\Delta w_{min}$ the convergence is poor since it controls the standard deviation of the stochastic update rule, for smaller $\Delta w_{min}$ the results are approaching the baseline model. The $\Delta w_{min}$ smaller than 0.01 produces an error penalty at the end of 30$^{th}$ epoch as small as just 0.3% above the 2.0% classification error of the baseline model.

To determine minimum and maximum conductance values that RPU devices should support for the algorithm to converge, a set of training curves is calculated as shown in Fig. 3B. Each curve is defined by the weight range where the absolute value of weights $|w_{ij}|$ is kept below a certain bound that is varied between 0.1 and 3. The other parameters are identical to Fig. 3A, while $\Delta w_{min}$ is taken as 0.001 to assure that the results are mostly defined by the choice of the weight range. The model with weights $|w_{ij}|$ bounded to values larger than 0.3 results in an acceptable error penalty criteria of 0.3% as defined above. Since, the parameter $\Delta g_{min}$ (and $g_{ij}$) is proportional to $\Delta w_{min}$ (and $w_{ij}$) through the amplification factor defined by peripheral circuitry, the number of coincidence events required to move the RPU device from its minimum to its maximum conductance value can be derived as $(\max(g_{ij}) - \min(g_{ij}))/\Delta g_{min} = (\max(w_{ij}) - \min(w_{ij}))/\Delta w_{min}$. This gives a lower estimate for the number of states that are required to be stored on an RPU device as 600.

In order to determine the tolerance of the algorithm to the variation in the incremental conductance change due to a single coincidence event $\Delta g_{min}$, the $\Delta w_{min}$ value used for each coincidence event is assumed to be a random variable with a Gaussian distribution. Corresponding results are shown in Fig. 3C, where the standard deviation is varied while the average $\Delta w_{min}$ value is set to 0.001. As it is seen, the algorithm is robust against the randomness on the weight change for each coincidence event and models with a standard deviation below 150% of the mean value reach acceptable 0.3% error penalty.

For stochastic models illustrated in Fig. 3D, yet another randomness, a device-to-device variation in the incremental conductance change due to a single coincidence event $\Delta g_{min}$, is introduced. In this case the $\Delta w_{min}$ used for each RPU device is sampled from a Gaussian distribution at the beginning of the training and then this fixed value is used throughout the training for each coincidence event. For all stochastic models in Fig. 3D, the average $\Delta w_{min}$ value of 0.001 is used while the standard deviation is varied for each model. Results show that the algorithm is also robust against the device-to-device variation and an acceptable error penalty can be achieved for models with a standard deviation up to 110% of the mean value.

To determine tolerance of the algorithm to the device-to-device variation in the upper and lower bounds of the conductance value, we assumed upper and lower bounds that are different for each RPU device for the models in Fig. 3E. The bounds used for each RPU device are sampled from a Gaussian distribution at the beginning of the training and are used throughout the training. For all of the stochastic models in Fig. 3E, mean value of 1.0 for upper bound (and $-1.0$ for lower bound) is used to assure that the results are mostly defined by the device-to-device variation in the upper and lower bounds. Fig. 3E shows that the algorithm is robust against the variation in the bounds and models with a standard deviation up to 80% of the mean can achieve acceptable 0.3% error penalty.

Fabricated RPU devices may also show different amounts of change in the conductance value due to positive ($\Delta g_{min}^+$) and negative ($\Delta g_{min}^-$) pulses as illustrated in Figs. 1C and 1D. To determine how much asymmetry between up and down changes the algorithm can tolerate, the up ($\Delta w_{min}^+$) and down ($\Delta w_{min}^-$) changes in the weight value are varied as shown in Figs. 3F and 3G. In both cases this global asymmetry is considered to be uniform throughout the whole RPU device array. For each model in Fig. 3F $\Delta w_{min}^+$ is fixed to 0.001 while $\Delta w_{min}^-$ is varied from 0.95 to 0.25 weaker than the up value. Similarly, Fig. 3G shows an analogous results for $\Delta w_{min}^-$ fixed to 0.001 while $\Delta w_{min}^+$ is varied. Results show that up and down changes need to be significantly balanced (within 5% of the mean) in order for the stochastic model to achieve an acceptable 0.3% error penalty.

In order to determine tolerance of the algorithm to the device-to-device variation in asymmetry, as opposed to a global asymmetry considered in Figs. 3F and 3G, the curves in Fig. 3H are calculated for various values of the standard deviation of $\Delta w_{min}^+/\Delta w_{min}^-$. The parameters $\Delta w_{min}^+$ and $\Delta w_{min}^-$ for each RPU device are sampled from a Gaussian distribution at the beginning of the training and then used throughout the training for each coincidence event. All the models assume that the average value of $\Delta w_{min}^+$ and $\Delta w_{min}^-$ is 0.001. The standard deviation of $\Delta w_{min}^+/\Delta w_{min}^-$ needs to be less than 6% of the mean value to achieve an acceptable 0.3% error penalty.

Analog computation is sensitive to various noise sources such as thermal noise, shot noise, etc that are all additive and can be modelled as a single unbiased Gaussian noise. Influence of noise penalty during the weight update cycle is already considered in Figs. 3C-3H. In order to estimate tolerance of the algorithm to noise during forward and backward cycles, we modelled analog noise as a random error imposed on the results of vector-matrix multiplication. As it is shown in Fig. 3I, an acceptable 0.3% error penalty is reached for a noise level of 10% normalized on activation function temperature.

Radar diagram in Fig. 4A summarizes specifications of RPU devices that are derived from the "stress tests" performed in Fig. 3. Axes C-I correspond to experiments in Figs. 3C-3I, respectively. Solid line 1 connects threshold values determined for these parameters for an acceptable 0.3% error penalty. Note that these specifications differ significantly from parameters typical for NVM technologies. The storage in NVM devices is digital and typically does not exceed a few bits. This constraint is imposed by system requirement to achieve high signal-to-noise ratio for read and write operations. In addition, the write operation does not depend on history as it overwrites all previously stored values. In contrast, weight values in the neural network operation are not needed to be written and resolved with very high signal-to-noise ratio. In fact, the algorithm can withstand up to 150% of noise in the weights updates (parameter C) and can tolerate up to 10% reading noise on columns or rows (parameter I). However, as opposed to a few bit storage capacity on NVM devices, a large number of coincidence events (over 600 from Fig. 3B) is required for the RPU device to keep track of the history of weight updates. In addition, in contrast to high endurance of full swing writing between bit levels required for NVM devices, RPU devices need to have high endurance only to small incremental changes, $\Delta g_{min}$.

Combined contribution of all parameters considered in Fig. 4A can be additive and therefore exceed the acceptable 0.3% error penalty. Fig. 4B shows training results when effects of more than one parameter are combined. When all parameters (C, D, E, F, G, H, and I) are combined at the threshold the test error reaches 5.0% that is 3.0% above the baseline model. Although this penalty can be acceptable for some applications, it is significantly higher than the 0.3% error penalty considered above.

This 3.0% penalty is higher than a simple additive impact of uncorrelated contributions indicating that at least some of these parameters are interacting. It opens the possibility of optimizing the error penalty by trading off tolerances between various parameters. For example, the model that combines only parameters C, D, and E at the threshold, as shown by curve 2 in Fig. 4B, gives 0.9% error penalty that is about the expected sum of individual contributions. Note that these parameters are defined by imperfections in device operation and by device-to-device mismatch that are all controlled by fabrication tolerances in a given technology. Even for deeply scaled CMOS technologies the fabrication tolerances do not exceed 30% that is much smaller than 150%, 110%, and 80% used for calculation of curve 2. The contributions of C, D and E to the error penalty can be eliminated by setting the corresponding tolerances to 30% (data not shown).

Among the parameters of Fig. 4A, the asymmetry between up and down changes in the conductance value of RPU devices (parameter F, G and H) is the most restrictive. Parameter F (or G) is the global asymmetry that can be compensated by controlling pulse voltages and/or number of pulses in the positive and negative update cycles, and hence even asymmetries higher than the threshold value of 5% can be eliminated with proper design of peripheral circuits. In contrast, the parameter H that is defined by device-to-device variation in the asymmetry, can be compensated by peripheral circuits only if each RPU device is addressed serially. To maintain the $O(1)$ time complexity, the device mismatch parameter H and the noise parameter I can be co-optimized to reduce the error penalty. The resulting model illustrated by the gray shaded area bounded with curve 3 in Fig. 4B achieves at most 0.3% error penalty. For this model parameters C, D, and E are set to 30% while F (or G) is set to zero, H is set to 2%, and I is set to 5%. Alternatively, the same result (data not shown) can be obtained by restricting the noise parameter I to 2.5% and increasing the device mismatch tolerance H to 4% that can simplify the array fabrication in expense of designing less noisy circuits.

5. **Circuit and system level design considerations**

The ultimate acceleration of DNN training with backpropagation algorithm on a RPU array of size $N \times N$ can be approached when $O(1)$ time complexity operation is enforced. In this case overall acceleration is proportional to $N^2$ that favors very large arrays. In general the design of the array, peripheral circuits, and the whole system should be based on optimization of the network parameters for a specific workload and classification task. In order to develop a general methodology for such a design, we will use the results of the analysis presented above as an example with understanding, however, that the developed approach is valid for larger class of more complicated cases than a relatively simple 3 layer network used to classify the MNIST dataset in Figs. 2-4.

### 5.1 RPU array design

For realistic technological implementations of the crossbar array, the array size will ultimately be limited by resistance and parasitic capacitance of the transmission lines resulting in significant $RC$ delay and voltage drop. For further analysis we assume that RPU devices are integrated at the back-end-of-line (BEOL) stack in-between intermediate metal levels. This allows the top thick metal levels to be used for power distribution, and the lower metal levels and the area under the RPU array for peripheral CMOS circuitry. Typical intermediate metal levels in a scaled CMOS technology have a thickness of 360 $nm$, and a width of 200 $nm$. Corresponding typical line resistance is about $r_{line} = 0.36\ \Omega/\mu m$ with parasitic capacitance of $c_{line} = 0.2\ fF/\mu m$. Assuming a reasonable 1 $GHz$ clock frequency for the pulses used during the update cycle, and allowing $RC$ delay to be at most 10% of the pulse width (0.1 $ns$), the longest line length should be $l_{line} = 1.64\ mm$. Assuming a reasonable line spacing of 200 $nm$ this results in an array with 4096 × 4096 RPU devices. Since the conductance values of RPU devices can only be positive, we assume that a pair of identical RPU device arrays is used to encode positive ($g_{ij}^+$) and negative ($g_{ij}^-$) weight values. The weight value ($w_{ij}$) is proportional to a difference of two conductance values stored in two corresponding devices ($g_{ij}^+ - g_{ij}^-$) located in identical positions of a pair of RPU arrays. To minimize the area, these two arrays can be stacked on top of each other occupying 4 consecutive metal levels resulting in a total area of $A_{array} = 2.68\ mm^2$. For this array size a full update cycle (both positive and negative) performed using 1 $ns$ pulses can be completed in 20 $ns$ for $BL = 10$.

In order to estimate an average RPU device resistance, $R_{device}$, we assume at most 10% voltage drop on the transmission line that is defined by $N \times R_{line}/R_{device}$, where $R_{line}$ is the total line resistance equal to $r_{line}l_{line}$. The contribution of the output resistance of the line drivers to the total line resistance can be minimized by proper circuit design. For an array size of $N = 4096$ the average RPU device resistance is therefore $R_{device} = 24\ M\Omega$. Using this resistance value, and assuming an operating voltage of 1 $V$ for all 3 training cycles and on-average about 20% activity for each device that is typical for the models of Figs. 2-4, the power dissipation on a pair of RPU arrays can be estimated as $P_{array} = 0.28\ W$.

### 5.2 Design of peripheral circuits

Operation of a single column (or row) during forward (or backward) cycle is illustrated in Fig. 5A. In contrast to the update cycle, stochastic translators are not needed. Here we assume that time-encoding scheme is used when input vectors are represented by fixed amplitude $V_{in} = 1\ V$ pulses with a tunable duration. Pulse duration is a multiple of 1 $ns$ and is proportional to the value of the input vector. Currents generated at each RPU device are summed on the columns (or rows) and this total current is integrated over the measurement time, $t_{meas}$ by current readout circuits as illustrated in Fig. 5A. Positive and negative voltage pulses are supplied separately to each of the identical RPU arrays that are used to encode positive and negative weights. Currents from both arrays are fed into peripheral circuitry that consists of an op-amp that integrates differential current on the capacitor $C_{int}$, and an analog-to-digital converter

ADC. Note, that for time-encoded pulses, the time-quantization error at the input to the RPU array scales inversely with the total number of pulses. For the models in Fig. 4B number of pulses larger than 20 (~5 bit resolution) is enough to eliminate corresponding error penalty.

We define a single RPU tile as a pair of arrays with $4096 \times 4096$ devices with peripheral circuits that support the parallel operation of the array in all 3 cycles. Peripheral circuitry includes ADCs, op-amps, STRs consisting of random number generators, and line drivers used to direct signals along the columns and rows. As shown in Fig. 5C the signals from a tile are directed towards a non-linear function (NLF) circuit that calculates either activation functions (i.e. sigmoid, softmax) and their derivates as well as arithmetical operations (i.e. multiplication) depending on cycle type and on position of corresponding layer. At the tile boundary input signals to the NLF are bounded to a certain threshold value to avoid signal saturation. Fig. 5B shows test error for the network of the model 3 in Fig. 4B, but with bounds $|\alpha|$ imposed on results of vector-matrix multiplication that is equivalent to restricting the NLF input. For neurons in hidden layers the NLF circuit should compute a sigmoid activation function. When the input to this sigmoid NLF is restricted to $|\alpha| = 3$, the resulting error penalty does not exceed an additional 0.4% as shown by curve 1 in Fig. 5B.

Neurons at the output layer perform a softmax NLF operation, that, when corresponding input is also restricted to $|\alpha| = 3$, results in exceedingly large error as shown by curve 2 in Fig. 5B. To make design more flexible and programmable it is desired for the NLF in both hidden and output layers to have the same bounds. When bounds on both softmax and sigmoid NLF are restricted to $|\alpha| = 12$, the total penalty is within acceptable range as shown by curve 3. Assuming 5% acceptable noise level taken from the results of Fig. 4B and an operation voltage range between $-1\,V$ and $1\,V$ at the input to the ADC, the corresponding bit resolution and voltage step required are $9\,bits$ and $4\,mV$, respectively. These numbers imply that the acceptable total integrated RMS voltage noise at the input to the ADC (or at the output of the op-amp) should not exceed $4\,mV$.

### 5.3 Noise analysis

In order to estimate the acceptable level of the input referred noise the integration function of the op-amp should be defined. Voltage at the output of the op-amp can be derived as

$$V_{out} = 2N \frac{V_{in}\, t_{meas}}{R_{device}\, C_{int}} \left(\frac{\beta - 1}{\beta + 1}\right) \qquad (4)$$

where $\beta$ is the conductance on/off ratio for an RPU device. This equation assumes all $N$ devices are contributing simultaneously that makes it hard to design a circuit that would require either a very large capacitor or large voltage swing. However, for a given bounds $|\alpha|$ imposed on the NLF transformation, the output voltage should not necessarily exceed the level corresponding to simultaneous contribution of $|\alpha|$ devices. Since, as shown above, an acceptable bound $|\alpha| = 12$ is enough, the number $N$ in Eq.4 can be replaced with 12. Assuming that $V_{out}$ signal feeding into the ADC should not exceed $1\,V$, and the $R_{device}$ is $24\,M\Omega$, the choice of integrating capacitor $C_{int}$ is dictated by the integration time $t_{meas}$ and on/off ratio $\beta$. Fig. 5D presents estimates of acceptable noise levels for various on/off ratios on the devices $\beta$ and integration times $t_{meas}$. This noise level corresponds to the input referred noise of the op-amp calculated using standard noise analysis in integrator-based circuits [35]. If $t_{meas}$ is taken as $20\ ns$ following the quantization error consideration discussed above, the acceptable noise levels are relatively low of the order of just $10\ nV/\sqrt{Hz}$ as seen in Fig. 5D curve 1. Even an increase of the on/off ratio $\beta$ to several orders of magnitude does not help to accommodate higher noise. In order to accommodate higher noise $t_{meas}$ needs to be increased with a penalty, however, of increased overall calculation time. As seen from curves in Fig. 5D, for a given noise level the on/off ratios as small as 2 to 10 can be acceptable that is, in fact, quite modest in comparison to several orders of magnitude typical for NVM applications. When $t_{meas}$ and $\beta$ are chosen as $80\ ns$ and 6, respectively, corresponding level of acceptable input referred noise shown by curve 2 in Fig. 5D can be derived as $15.1\ nV/\sqrt{Hz}$. Corresponding capacitance $C_{int}$ can also be calculated as $57\ fF$ using Eq. 4.

Various noise sources can contribute to total acceptable input referred noise level of an op-amp including thermal noise, shot noise, and supply voltage noise, etc. Thermal noise due to a pair of arrays with $4096 \times 4096$ RPU devices can be estimated as $7.0\ nV/\sqrt{Hz}$, which leaves about $13.4\ nV/\sqrt{Hz}$ for other noise sources. Depending on exact physical implementation of a RPU device and type of non-linear $I - V$ response, shot noise levels produced by the RPU array can vary. Assuming a diode-like model, total noise from a whole array scales as a square root of a number of active RPU devices in a column (or a row), and hence depends on an overall instantaneous activity of a network. For a pair of arrays with $4096 \times 4096$ RPU devices and assuming a moderate 20% activity of the network that is typical for the models of Figs. 2-4, the shot noise contribution is about $13.4\ nV/\sqrt{Hz}$. Longer integration time is needed for higher workloads or additional noise contributions including the noise on the voltage, amplifier noise, etc.

### 5.4 System level design considerations

The tile area occupied by peripheral circuitry and corresponding dissipated power are dominated by the contribution from 4096 ADC. Assuming $t_{meas}$ of $80\ ns$ for forward and backward cycles, ADCs operating with $9\ bit$ resolution at $12.5\ MSamples/sec$ are required. The state-of-the-art SAR-ADC [36,37] that can provide this performance, occupy an area of $0.0256\ mm^2$ and consume $0.24\ mW$ power that results in a total area of $104\ mm^2$ and a total power of $1\ W$ for an array of 4096 ADCs. This area is much larger than the RPU array itself, therefore it is reasonable to time-multiplex the ADCs between different columns/rows by increasing the sampling rate while keeping total power unchanged. Assuming each ADC is shared by 64 columns (or rows), the total ADC area can be reduced to $1.64\ mm^2$ with each ADC running at about $800\ MSamples/sec$. Since we assume that RPU device arrays are built on the intermediate metal levels on top of peripheral CMOS circuitry, the total tile area is defined by the RPU array area of $2.68\ mm^2$ that leaves about $1.0\ mm^2$ for other circuitry that also can be area optimized. For example, the number of random number generators used to translate binary data to stochastic bit stream can be significantly reduced to just 2 as no operations are performed on streams generated within columns (or rows) and evidenced by no additional error penalty for corresponding classification test (data not shown). Total area of a single tile therefore is $2.68\ mm^2$, while the total power dissipated by both RPU arrays and all peripheral circuitry (ADCs, opamps, STR) can be estimated as $2.0\ W$, assuming $0.7\ W$ reserved for op-amps and STRs.

The number of weight updates per second on a single tile can be estimated as 839 $TeraUpdates/s$ given the 20 $ns$ duration of the update cycle and 4096 × 4096 array size. This translates into power efficiency of 419 $TeraUpdates/s/W$ and area efficiency of 319 $TeraUpdates/s/mm^2$. The tile throughput during the forward and backward cycles can be estimated as 419 $TeraOps/s$ given 80 $ns$ for forward (or backward) cycle with power and area efficiencies of 210 $TeraOps/s/W$ and 156 $TeraOps/s/mm^2$, respectively. These efficiency numbers are about 5 orders of magnitude better than state-of-the-art CPU and GPU performance metrics [38].

The power and area efficiencies achieved for a single tile will inevitably degrade as multiple tiles are integrated together as a system-on-chip. As illustrated in Fig. 5C, additional power and area should be reserved for programmable NLF circuits, on-chip communication via coherent bus or network-on-chip (NoC), off-chip I/O circuitry, etc. Increasing the number of tiles on a chip will first result in an acceleration of a total chip throughput, but eventually would saturate as it will be limited either by power, area, communication bandwidth or compute resources. State-of-the-art high-performance CPU (IBM Power8 12-core CPU [39]) or GPU (NVidia Tesla K40 GPU [40]) can be taken as a reference for estimation of the maximum area of 600 $mm^2$ and power of 250 $W$ on a single chip. While power and area per tile are not prohibitive to scale the number of tiles up to 50 to 100, the communication bandwidth and compute resources needed for a system to be efficient might be challenging.

Communication bandwidth for a single tile can be estimated assuming 5 bit input and 9 bit output per column (or row) for forward (or backward) cycles that give in total about 90 GB/s unidirectional bandwidths that will also satisfy the update cycle communication requirements. This number is about 3 times less than the communication bandwidth in IBM Power8 CPU between a single core and a nearby L2 cache [39]. State-of-the-art on-chip coherent bus (over 3 TB/s in IBM Power8 CPU [39]) or NoC (2.5 TB/s in Ref [41]) can provide sufficient communication bandwidth between distant tiles.

Compute resources needed to sustain $O(1)$ time complexity for a single tile can be estimated as 51 $GigaOps/s$ assuming 80 $ns$ cycle time and 4096 numbers generated at columns or rows. To support parallel operation of $n$ tiles, compute resources need to be scaled by $O(n)$ thus limiting the number of tiles that can be active at a given time to keep the total power envelop on a chip below 250 $W$. For example, a single core of IBM Power8 CPU [39] can achieve about 50 $GigaFLOP/s$ that might be sufficient to support one tile, however the maximum power is reached just for 12 tiles assuming 20 W per core. Corresponding power efficiency for this design point (Design 1 in Table 1) would be 20 $TeraOps/s/W$. Same compute resources can be provided by 32 cores of state-of-the-art GPU [40], but with better power efficiency thus allowing up to 50 tiles to work in parallel. Corresponding power efficiency for this design (Design 2 in Table 1) would be 84 $TeraOps/s/W$. Further increase in the number of tiles that can operate concurrently can be envisioned by designing specialized power and area efficient digital circuits that operate fixed point numbers with limited bit resolution. An alternative design (Design 3 in Table 1) can be based on just a few compute cores that can process the tile data sequentially in order to fit larger numbers of tiles to deal with larger network sizes. For example, a chip with 100 tiles and a single 50 $GigaOps/s$ compute core will be capable of dealing with networks with as many as 1.6 billion weights and dissipate only about 22 W assuming 20W from compute core and communication bus and just 2 W for RPU tiles since only one is active at any given time. This gives a power efficiency of $20\ TeraOps/s/W$ that is 4 orders of magnitude better than state-of-the-art CPU and GPU.

## 6. Discussion

We proposed a concept of resistive processing unit (RPU) devices that can simultaneously store and process data locally and in parallel, thus potentially providing significant acceleration for DNN training. The tolerance of the training algorithm to various RPU device and system parameters as well as to technological imperfections and different sources of noise has been explored. This analysis allows to define a list of specifications for RPU devices summarized in Table 2 that can be used as a guide for a systematic search for new physical mechanisms, materials and device designs to realize the RPU device concept with realistic CMOS-compatible technology.

We also presented an analysis of various system designs based on the RPU array concept that can potentially provide many orders of magnitude acceleration of deep neural network training while significantly decreasing required power and compute hardware resources. The results are summarized in Table 1. This analysis shows that, depending on the network size, different design choices for the RPU accelerator chip can be made that trade power and acceleration factor.

The proposed accelerator chip design of Fig. 5C is flexible and can accommodate different types of DNN architectures beyond fully connected layers. For example, convolutional layers can be also mapped to an RPU array in an analogous way. In this case, instead of performing a vector-matrix multiplication for forward and backward cycles, an array needs to perform a matrix-matrix multiplication that can be achieved by feeding the columns of the input matrix serially into the columns of the RPU array. In addition, peripheral NLF circuits need to be reprogrammed to perform not only calculation of activation functions, but also max-pooling and sub-sampling. The update cycle operations are identical for both convolutional and fully connected layers hence do not require reprograming. The required connectivity between layers can be achieved by reprogramming tile addresses in a network.

Most of the recent DNN architectures are based on combination of many convolutional and fully connected layers [ref ref] with a number of parameters of the order of a billion. Our analysis demonstrates that a single RPU accelerator chip can be used to train such a large deep neural networks. Problems of the size of ImageNet classification that currently require days of training on a datacenter-size cluster with thousands of machines [6] can take just a few hours on a single RPU accelerator chip.

# References

1. LeCun, Y., Bengio, Y. & Hinton, G. Deep learning. *Nature* **521,** 436–444 (2015).
2. Hinton, G *et al.* Deep neural networks for acoustic modeling in speech recognition: The shared views of four research groups. *IEEE Signal Processing Magazine* **29,** 82–97 (2012).
3. Krizhevsky, A, Sutskever, I & Hinton, GE. Imagenet classification with deep convolutional neural networks. *NIPS* 1097–1105 (2012).
4. Simonyan, K. & Zisserman, A. Very Deep Convolutional Networks for Large-Scale Image Recognition. *ICLR* (2015).
5. Szegedy, C. *et al.* Going Deeper with Convolutions. *CVPR* (2015).
6. Le, Q. *et al.* Building high-level features using large scale unsupervised learning. *International Conference on Machine Learning* (2012).
7. Rumelhart, DE, Hinton, GE & Willims, RJ. Learning representations by back-propagating errors. *Nature* **323,** 533–536 (1986).
8. Lehmann, C, Viredaz, M & Blayo, F. A generic systolic array building block for neural networks with on-chip learning. *IEEE Transactions on Neural Networks* **4,** 400–407 (1993).
9. Arima, Y *et al.* A 336-neuron, 28 K-synapse, self-learning neural network chip with branch-neuron-unit architecture. *IEEE Journal of Solid-State Circuits* **26,** 1637–1644 (1991).
10. Coates, A *et al.* Deep learning with COTS HPC systems. *ICML* (2013).
11. Wu, R., Yan, S., Shan, Y., Dang, Q. & Sun, G. Deep Image: Scaling up Image Recognition. *arXiv:1501.02876 [cs.CV]* (2015).
12. Gupta, S., Agrawal, A., Gopalakrishnan, K. & Narayanan, P. Deep Learning with Limited Numerical Precision. *arXiv:1502.02551 [cs.LG]* (2015).
13. Chen, Y, Luo, T, Liu, S, Zhang, S & He, L. DaDianNao: A Machine-Learning Supercomputer. *2014 47th Annual IEEE/ACM International Symposium on Microarchitecture* 609–622 (2014). doi:10.1109/MICRO.2014.58
14. Jackson, B. *et al.* Nanoscale electronic synapses using phase change devices. *Acm J. Emerg. Technologies Comput. Syst.* **9,** 1–20 (2013).
15. Kuzum, D., Yu, S. & Wong, H.-S. P. S. Synaptic electronics: materials, devices and applications. *Nanotechnology* **24,** 382001 (2013).
16. Yu, S. *et al.* A Low Energy Oxide Based Electronic Synaptic Device for Neuromorphic Visual Systems with Tolerance to Device Variation. *Advanced Materials* **25,** 1774–1779 (2013).
17. Saïghi, S. *et al.* Plasticity in memristive devices for spiking neural networks. *Frontiers in Neuroscience* **9,** (2015).
18. Jo, S. *et al.* Nanoscale memristor device as synapse in neuromorphic systems. *Nano letters* **10,** 1297–301 (2010).
19. Indiveri, G., Linares-Barranco, B., Legenstein, R., Deligeorgis, G. & Prodromakis, T. Integration of nanoscale memristor synapses in neuromorphic computing architectures. *Nanotechnology* **24,** 384010 (2013).
20. Bi, G. Q. & Poo, M. M. Synaptic modifications in cultured hippocampal neurons: dependence on spike timing, synaptic strength, and postsynaptic cell type. *The Journal of neuroscience : the official journal of the Society for Neuroscience* **18,** 10464–10472 (1998).
21. Xu, Z. *et al.* Parallel Programming of Resistive Cross-point Array for Synaptic Plasticity. *Procedia Computer Science* **41,** 126–133 (2014).
22. Burr, GW *et al.* Experimental demonstration and tolerancing of a large-scale neural network (165,000 synapses), using phase-change memory as the synaptic weight element. *Electron Devices Meeting (IEDM), 2014 IEEE International* 29.5.1 (2014). doi:10.1109/IEDM.2014.7047135
23. Li, B., Wang, Y., Wang, Y., Chen, Y. & Yang, H. Training itself: Mixed-signal training acceleration for memristor-based neural network. 361–366 (2014).

24. Soudry, D., Castro, D., Gal, A., Kolodny, A. & Kvatinsky, S. Memristor-Based Multilayer Neural Networks With Online Gradient Descent Training. *IEEE Transactions on Neural Networks and Learning Systems* **26,** 2408–2421 (2015).
25. Prezioso, M *et al.* Training and operation of an integrated neuromorphic network based on metal-oxide memristors. *Nature* **521,** 61–64 (2015).
26. Burr, GW *et al.* Largescale neural networks implemented with nonvolatile memory as the synaptic weight element: comparative performance analysis (accuracy, speed, and power). *IEDM (International Electron Devices Meeting)* (2015).
27. Seo, J. *et al.* On-Chip Sparse Learning Acceleration With CMOS and Resistive Synaptic Devices. *IEEE Transactions on Nanotechnology* **14,** 969–979 (2015).
28. Yu, S *et al.* Scaling-up Resistive Synaptic Arrays for Neuro-inspired Architecture: Challenges and Prospect. *IEDM* 451–454 (2015).
29. Steinbuch, K. Die Lernmatrix. *Kybernetik* **1,** 36–45 (1961).
30. Poppelbaum, WJ, Afuso, C & Esch, JW. Stochastic computing elements and systems. *In Proceedings of the AFIPS Fall Joint Computer Conference* 635–644 (1967).
31. Gaines, BR. Stochastic computing. *In Proceedings of the AFIPS Spring Joint Computer Conference* 149–156
32. Alaghi, A. & Hayes, J. Survey of Stochastic Computing. *Acm Transactions Embed. Comput. Syst.* **12,** 1–19 (2013).
33. LeCun, Y, Bottou, L, Bengio, Y & Haffner, P. Gradient-based learning applied to document recognition. *Proceedings of the IEEE* **86,** 2278–2324
34. Strukov, D., Snider, G., Stewart, D. & Williams, R. The missing memristor found. *Nature* **453,** 80–83 (2008).
35. Jensen, K, Gaudet, VC & Levine, PM. Noise analysis and measurement of integrator-based sensor interface circuits for fluorescence detection in lab-on-a-chip applications. *ICNF* (2013). doi:10.1109/ICNF.2013.6578905
36. Jonsson, BE. An empirical approach to finding energy efficient ADC architectures. *2011 International Workshop on ADC Modelling, Testing and Data Converter Analysis and Design and IEEE 2011 ADC Forum* 132–137 (2011).
37. Jonsson, BE. Area Efficiency of ADC Architectures. *2011 20th European Conference on Circuit Theory and Design (ECCTD)* 560–563 (2011).
38. Gokhale, V, Jin, J, Dundar, A & Martini, B. A 240 g-ops/s mobile coprocessor for deep neural networks. *IEEE Conference on Computer Vision and Pattern Recognition Workshops (CVPRW)* 696–701 (2014). doi:10.1109/CVPRW.2014.106
39. Stuecheli, J. Next Generation POWER microprocessor. *Hot Chips Conference* (2013).
40. NVIDIA. NVIDIA's next generation CUDA compute architecture: Kepler GK110. Whitepaper, 2012.
41. Chen, G *et al.* A 340 mV-to-0.9 V 20.2 Tb/s Source-Synchronous Hybrid Packet/Circuit-Switched 16× 16 Network-on-Chip in 22 nm Tri-Gate CMOS. *2014 IEEE International Soild-State Circuits Conference Digest of techincal Papers (ISSCC)* 276 – 277 (2014). doi:10.1109/JSSC.2014.2369508

**Acknowledgments :** We thank Seyoung Kim, Jonathan Proesel, Mattia Rigotti, Wilfried Haensch, Geoffrey Burr, Michael Perrone, Bruce Elmegreen, Suyog Gupta, Ankur Agrawal, Mounir Meghelli, Dennis Newns, and Gerald Tesauro for many useful discussions and suggestions.

**Figures and Tables**

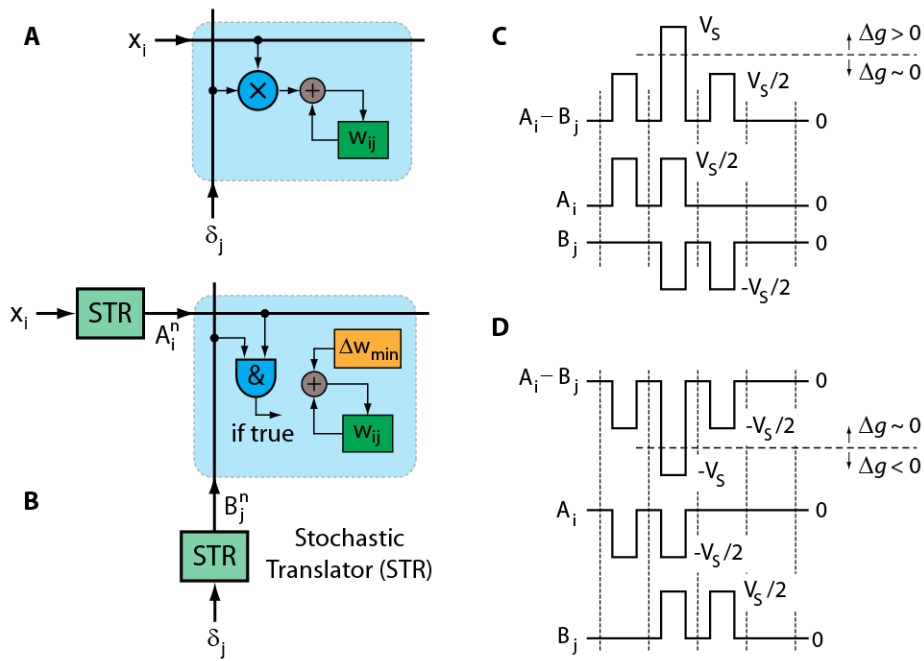

**Fig. 1.** (**A**) Schematics of original weight update rule of Eq.1 performed at each cross-point. (**B**) Schematics of stochastic update rule of Eq.2 that uses simple AND operation at each cross-point. Pulsing scheme that enables the implementation of stochastic updates rule by RPU devices for (**C**) up and (**D**) down conductance changes.

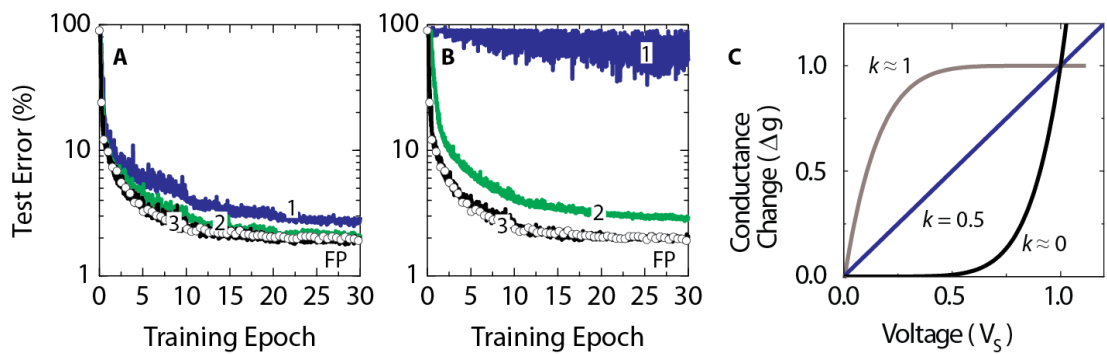

**Fig. 2.** Test error of DNN with the MNIST dataset. Open white circles correspond to the baseline model with the training performed using the conventional update rule of Eq.1. (**A**) Lines marked as 1, 2, and 3 correspond to the stochastic model of Eq.2 with stochastic bit lengths $BL = 1, 2$ and $10$, respectively. (**B**) Lines marked as 1, 2, and 3 correspond to the stochastic model with $BL = 10$ and the non-linearity ratio $k = 0.5, 0.4$ and $0.1$, respectively. (**C**) Illustration of various non-linear responses of RPU device with $k = 0, 0.5$ and $1$.

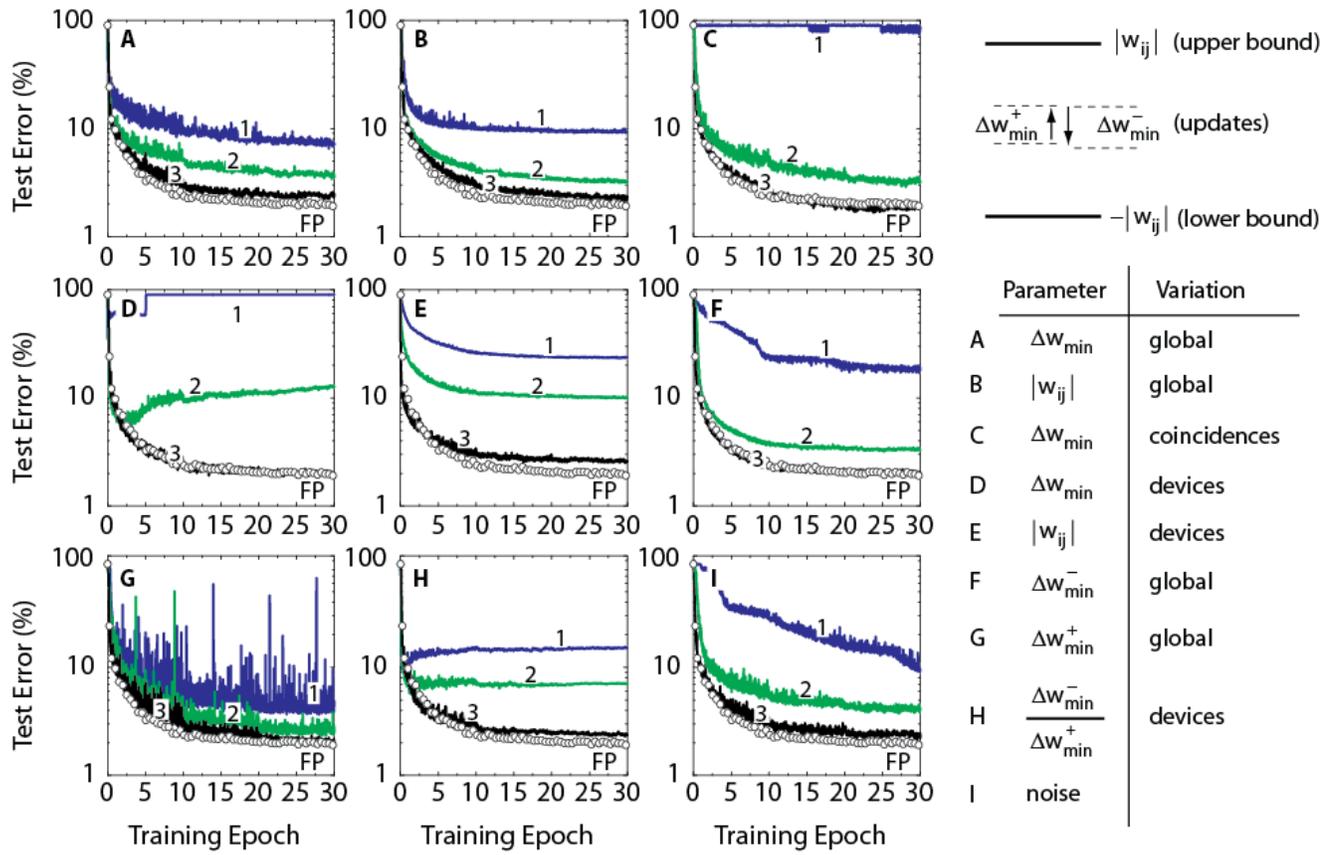

**Fig. 3.** Test error of DNN with the MNIST dataset. Open white circles correspond to a baseline model with the training is performed using the conventional update rule of Eq.1. All solid lines assume a stochastic model with $BL = 10$ and $k = 0$. (**A**) Lines 1, 2, and 3 correspond to a stochastic model with $\Delta w_{min} = 0.1, 0.032$ and $0.01$, respectively. All curves in B-I use $\Delta w_{min} = 0.001$. (**B**) Lines 1, 2, and 3 correspond to a stochastic model with weights bounded to 0.1, 0.2, and 0.3, respectively. (**C**) Lines 1, 2, and 3 correspond to a stochastic model with a coincidence-to-coincidence variation in $\Delta w_{min}$ of 1000%, 320%, and 100%, respectively. (**D**) Lines 1, 2, and 3 correspond to a stochastic model with device-to-device variation in $\Delta w_{min}$ of 1000%, 320%, and 100%, respectively. (**E**) Lines 1, 2, and 3 correspond to a stochastic model with device-to-device variation in the upper and lower bounds of 1000%, 320%, and 100%, respectively. All solid lines in E have a mean value of 1.0 for upper bound (and −1.0 for lower bound). (**F**) Lines 1, 2, and 3 correspond to a stochastic model, where down changes are weaker by 0.5, 0.75, and 0.9, respectively. (**G**) Lines 1, 2, and 3 correspond to a stochastic model, where up changes are weaker by 0.5, 0.75, and 0.9, respectively. (**H**) Lines 1, 2, and 3 correspond to a stochastic model with device-to-device variation in the up and down changes by 40%, 20%, and 6%, respectively. (**I**) Lines 1, 2, and 3 correspond to a stochastic model with a noise in vector-matrix multiplication of 100%, 60%, and 10%, respectively, normalized on activation function temperature.

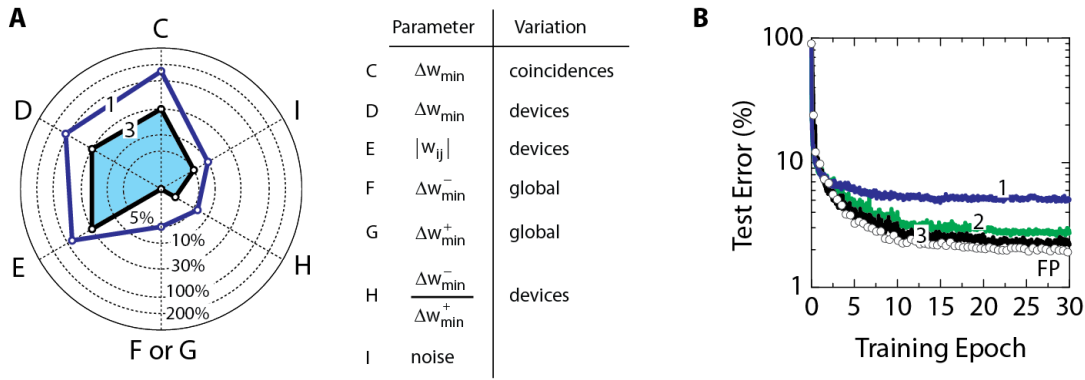

**Fig. 4.** (**A**) Line 1 shows threshold values for parameters from Fig. 3 assuming a 0.3% error penalty. Parameters C-I correspond to experiments in Figs. 3C-3I, respectively. The gray shaded area bounded by line 3 results in at most 0.3% error penalty when all parameters are combined. (**B**) Curve 1 corresponds to a model with all parameters combined at the threshold value as shown in the radar diagram by line 1. Curve 2 corresponds to a model with only C, D and E combined at the threshold. Curve 3 corresponds to a model with C, D, E at 30%, F/G at 0%, H at 2% and I at 5%, all combined as shown in the radar diagram by line 3.

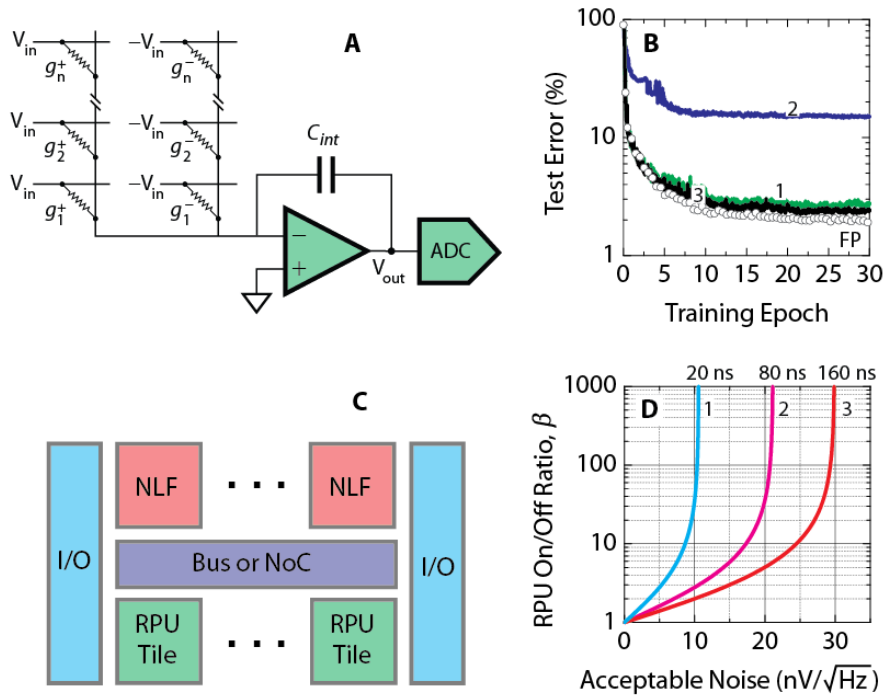

**Fig. 5.** (**A**) Operation of a single column (or row) during forward (or backward) cycle showing an op-amp that integrates the differential current on the capacitor $C_{int}$, and an analog-to-digital converter ADC. (**B**) Test error for the network of model 3 in Fig. 4B with bounds $|\alpha|$ imposed on results of vector-matrix multiplication. Curve 1 corresponds to a model with $|\alpha| = 3$ imposed only on sigmoid activation function in hidden layers. Curves 2 and 3 corresponds to a model with $|\alpha| = 3$ and 12, respectively, imposed on both sigmoid and softmax activation functions. (**C**) Schematics of the architecture for accelerator RPU chip. RPU tiles are located on the bottom, NLF digital compute circuits are on the top, on-chip communication is provided by a bus or NoC, and off-chip communication relies on I/O circuits. (**D**) Acceptable input referred noise levels for various on/off ratio on the RPU devices $\beta$ and integration times $t_{meas}$. Curves 1,2 and 3 corresponds to $t_{meas}$ of 20, 80 and 160 ns, respectively.

**Table 1. Summary of comparison of various RPU system designs and state-of-the-art CPU and GPU**

| System | Throughput (TeraOps/s) | Power (W) | Power Efficiency (G-Ops/s/W) | Network Size (Number of weights) | Acceleration vs CPU |
|---|---|---|---|---|---|
| CPU Power8 12 Cores | 0.676 | 250 | 2.7 | - | 1 |
| GPU NVidia Tesla K40 | 4.3 | 242 | 17.8 | - | 6.4 |
| Design 1 | 5,000 | 250 | 20,100 | 200 M | 7,400 |
| Design 2 | 21,000 | 250 | 83,800 | 840 M | 31,000 |
| Design 3 | 420 | 22 | 19,000 | 1,680 M | 620 |

**Table 2 : Summary of RPU device specifications**

| Specs | Parameter | Value | Tolerance |
|---|---|---|---|
| **Pulse Duration** |  | 1 ns |  |
| **Operating Voltage** | $\pm V_S$ | 1 V |  |
| **Maximum Device Area** |  | 0.04 $\mu m^2$ |  |
| **Average Device Resistance** | $R_{device}$ | 24 $M\Omega$ | 7 $M\Omega$ |
| **Maximum Device Resistance** | $\max(g_{ij})$ | 84 $M\Omega$ | 7 $M\Omega$ |
| **Minimum Device Resistance** | $\min(g_{ij})$ | 14 $M\Omega$ | 7 $M\Omega$ |
| **Resistance change at $\pm V_S$** | $\Delta g_{min}^{\pm}$ | 70 $K\Omega$ | 21 $K\Omega$ |
| **Resistance change at $\pm V_S/2$** |  | 7 $K\Omega$ |  |
| **Storage capacity** | $(\max(g_{ij}) - \min(g_{ij}))/\Delta g_{min}$ | 1000 levels |  |
| **Device up/down asymmetry*** | $\Delta g_{min}^{+}/\Delta g_{min}^{-}$ | 1.05 | 2% |

Note that these numbers are derived from the radar diagram in Fig 4A and correspond to the shaded area.

*Up/down asymmetry of averaged over all devices can be to a large extend compensated by proper adjustment of pulse widths and/or pulse amplitude.